\pdfoutput=1

\documentclass[11pt]{article}

\usepackage[final]{acl}

\usepackage{times}
\usepackage{latexsym}
\usepackage{amsmath} 
\usepackage{tabularray}
\usepackage{booktabs}
\usepackage{multirow}
\usepackage{colortbl}
\usepackage{xcolor}
\usepackage{color}
\usepackage[inline]{enumitem}
\usepackage{multicol}
\usepackage{pifont}
\usepackage{booktabs,caption}
\usepackage[flushleft]{threeparttable}
\usepackage[most]{tcolorbox}
\usepackage{inconsolata}
\usepackage{makecell}
\usepackage{amssymb}
\usepackage{tabularx} 
\usepackage{hyperref}
\usepackage{amsmath}
\usepackage[T1]{fontenc}

\usepackage[utf8]{inputenc}

\usepackage{microtype}

\usepackage{graphicx}

\title{Divide-Then-Aggregate: An Efficient Tool Learning Method \\ via Parallel Tool Invocation}

\author{
    Dongsheng Zhu\textsuperscript{\rm1}$^{\dag}$,
    Weixian Shi\textsuperscript{\rm1}$^{\dag}$,
    Zhengliang Shi\textsuperscript{\rm2 \textdaggerdbl}, 
    \\
    \textbf{Zhaochun Ren\textsuperscript{\rm3}},
    \textbf{Shuaiqiang Wang\textsuperscript{\rm1}}, 
    \textbf{Lingyong Yan\textsuperscript{\rm1}$^*$},
    \textbf{Dawei Yin\textsuperscript{\rm1}$^*$}
    \\
    \textsuperscript{1}Baidu Inc., Beijing, China \ \ \textsuperscript{2}Shandong University, Qingdao, China \\
    \textsuperscript{3}Leiden University, Leiden, The Netherlands \\
    \texttt{\{zhudongsheng, shiweixian, yanlingyong, yindawei\}@baidu.com} \\
    \texttt{zhengliang.shii@gmail.com}\\
}

\definecolor{backred}{RGB}{255, 190, 190}
\definecolor{backblue}{RGB}{220, 230, 250}

\newtcbox{\hlprimarytab}{on line, rounded corners, box align=base, colback=backblue, colframe=white, size=fbox, arc=3pt, before upper=\strut, top=-2pt, bottom=-4pt, left=-2pt, right=-2pt, boxrule=0pt}
\newtcbox{\hlsecondarytab}{on line, box align=base, colback=backred, colframe=white, size=fbox, arc=3pt, before upper=\strut, top=-2pt, bottom=-4pt, left=-2pt, right=-2pt, boxrule=0pt}

\definecolor{Gainsboro}{rgb}{0.86, 0.86, 0.86}
\definecolor{Gray}{gray}{0.95}
\definecolor{LightCyan}{rgb}{0.88,1,1}

\newcommand{\besttext}[1]{{\hlsecondarytab{#1}}}
\newcommand{\hightext}[1]{{\hlprimarytab{#1}}}

\newcommand{\best}{\cellcolor{backred}\textbf}
\newcommand{\high}{\cellcolor{backblue}\textbf}

\begin{document}
\maketitle

\def\thefootnote{$^{\dag}$}\footnotetext{Equal contribution.}
\def\thefootnote{\textdaggerdbl}\footnotetext{Work done during internship.}
\def\thefootnote{*}\footnotetext{Co-corresponding authors.}

\begin{abstract}

While Large Language Models (LLMs) demonstrate remarkable capabilities, their ability to autonomously execute complex real-world tasks remains limited. Accordingly, tool learning has emerged to enable LLMs to effectively leverage external tools to extend their capabilities.
Current tool-learning paradigms like CoT/ReAct employ sequential tool invocation but suffer from constrained perception and inadequate task planning. Alternative approaches using search-based decision trees incur substantial computational overhead. 
To address these limitations, we propose DTA-Llama (\textbf{D}ivide-\textbf{T}hen-\textbf{A}ggregate Llama), a novel parallel tool invocation framework featuring: (1) A Directed Acyclic Graph (DAG) structure that transformed from traditional tree-based tool search paths, enabling parallel execution and contributing high-quality training data; (2) A process-thread-inspired inference mechanism that iteratively decomposes tasks into parallel tool-using subtasks while aggregating results for subsequent decisions. 
Experimental results show that our approach substantially enhances task performance while reducing token consumption and inference time. Llama2-7B, using our method, is comparable to the official parallel function calling method of GPT-3.5. The relevant code, dataset, and model weights are available at \href{https://corn0205.github.io/}{https://corn0205.github.io/}.

\end{abstract}

\section{Introduction}

Large Language Models (LLMs), which are pre-trained and fine-tuned on massive amounts of textual data, have demonstrated powerful proficiency in various artificial intelligence tasks,  such as conversation \citep{zheng2023judging,zhu2024vislinginstruct}, logical reasoning \citep{pan2023logic} and coding \citep{nijkampcodegen}.
However, more real-world tasks often require the LLMs to interact with the environment to get necessary external information or feedback, such as checking real-time flight status~\citep{guan2024intelligent} and complicate calculation for data analysis~\citep{sun2024lambda}.
To this end, tool learning has emerged recently which aims to equip the LLMs with external tools and teach them how to leverage the tools to accomplish real-world tasks.

\begin{figure}[tp]
\centering
\includegraphics[width=\columnwidth]{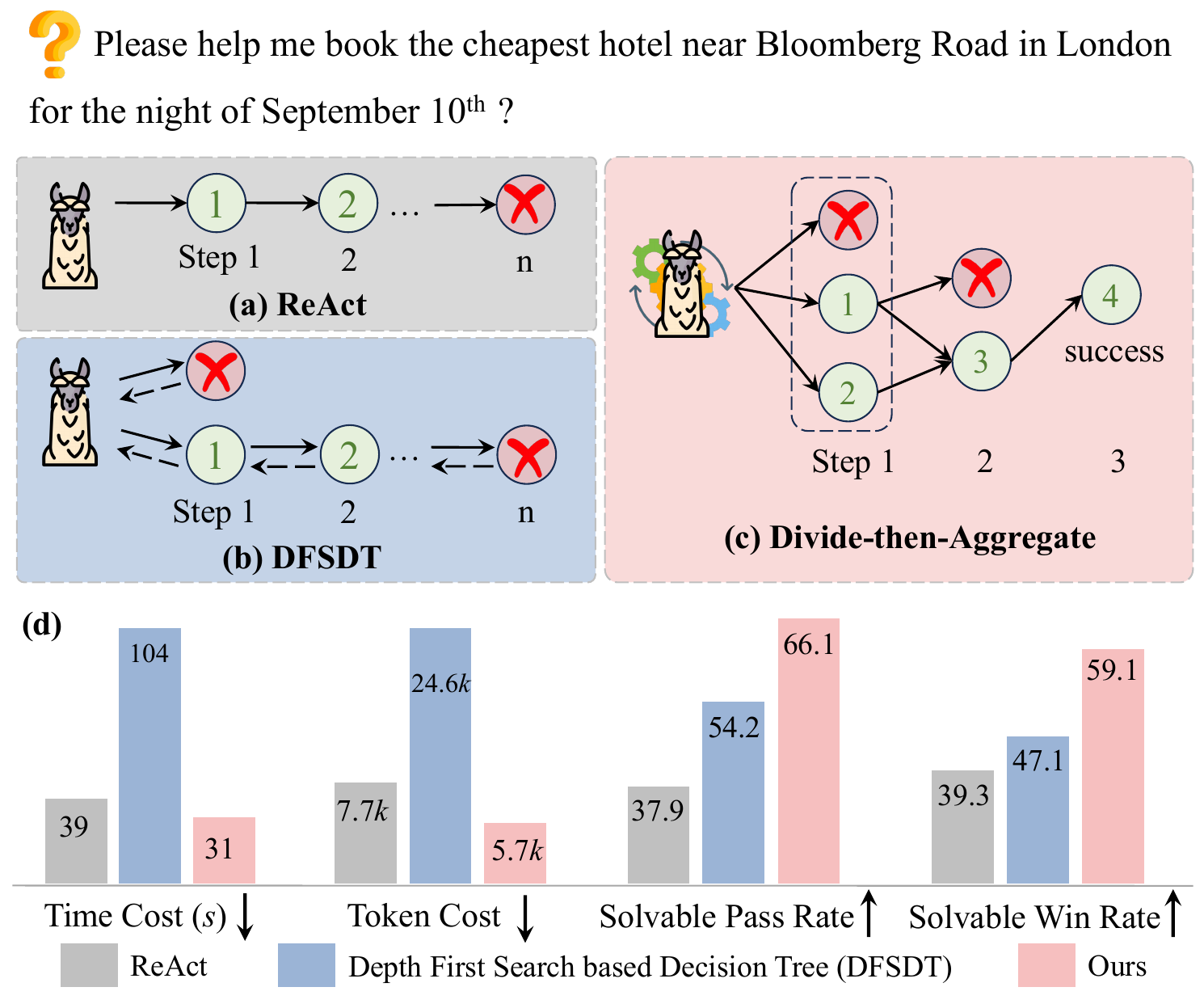}
\caption{
The top block depicts the comparison between CoT/ReAct, DFSDT and our proposed method. The bottom block provides the performance in four aspects of our method and baselines on benchmarks.
}
\label{fig:compare}
\end{figure}

Previous tool learning methods typically work in pipelined and tree-based paradigms.
Concretely, early studies usually perform the tool invocation in a pipeline, such as the Chain-of-Thought (CoT) reasoning \citep{kojima2022large}, and the ReAct mechanism \citep{yao2022ReAct}. In these methods, tool learning agents (i.e., the LLMs) interact with the environment through a \textit{Thought-Action-Observation} framework (as shown in Figure~\ref{fig:compare}(a)).
However, these methods usually focus on reflecting and planning based on local observations, rather than globally perceiving and planning the whole task solving paths.
In contrast, the tree-based methods, such as ToolLLM \citep{qin2023toolllm} and Toolchain*~\citep{zhuang2023toolchain}, adopt tree-based algorithms, e.g., Depth First Search based Decision Tree (DFSDT), to perform the global planning of tool invocation.
However, these methods still confront the inevitable backtracking mechanism that iteratively retries, which can usually significantly increase token consumption and inference time (as shown in Figure~\ref{fig:compare}(d) for an example).
Moreover, both two kinds of methods suffer from the limitation that LLMs invoke only one tool at each round during tool planning, which narrows their perceptual scope and necessitates more rounds of tool invocation, thereby reducing overall efficiency. 

In this paper, we introduce a novel model-based tool, \textbf{DTA-Llama} (\textbf{D}ivide-\textbf{T}hen-\textbf{A}ggregate Llama), which enables parallel tool invocation within each round of execution.
Specifically, we convert traditional tree-based tool search paths into a Directed Acyclic Graph (DAG) structure with level-order traversal (as illustrated in Figure~\ref{fig:compare}(c)), allowing for parallel execution of tools compared to previous sequential methods.
Using the widely adopted ToolBench dataset~\citep{qin2023toolllm}, we construct a high-quality parallel tool invocation dataset, \textbf{DTA-Tool}, and train Llama~\citep{touvron2023llama1,touvron2023llama,dubey2024llama} on it to develop DTA-Llama.
Additionally, we design a parallel tool invocation framework inspired by the Process/Threads mechanism~\citep{10.5555/1208561} for inference. In this framework, the \textbf{Process} component plans the tool invocation and divides parallelizable tools into separate \textbf{Threads}, which then execute independently according to the plan. After execution, an intermediate state lock aggregates the results from all threads.
This design shortens the invocation path and significantly improves the efficiency of tool use in LLMs.

We evaluate our approach on StableToolBench~\cite{guo2024stabletoolbench}, a comprehensive and reliable real-world tool-use benchmark. Performance is measured using solvable pass rate (SoPR), solvable win rate (SoWR), and actual computational cost.
Compared to existing methods, our approach achieves superior tool invocation performance while reducing computational cost—even matching GPT-3.5's~\citep{gpt3.5} function-calling performance using only a fine-tuned Llama2-7B~\citep{touvron2023llama}.
To further assess generalization, we fine-tune multiple LLMs, demonstrating the robustness and generalization ability of our method across different models.

In summary, our main contributions are as follows: 
\begin{itemize}
    \item We transform the tree-based serial data into a DAG format, contributing a high-quality and high-quantity parallel tool invocation dataset to the open-source community.
    \item A new tool invocation framework has been established, transforming invocation into the Process/Threads format. Combined with the parallel paradigm, this greatly simplifies the invocation path and improves efficiency.
    \item We comprehensively validate the superiority of DTA-Llama in real-world tool benchmarks, evaluating its performance from three aspects: effectiveness, computational cost, and generalization ability.
\end{itemize}

\section{Related Work}

\paragraph{\textbf{Tool Learning}}
The agent tool learning aims to expand LLMs capabilities by teaching LLMs to use external tools.
Many early studies~\cite{patil2023gorilla,tang2023toolalpaca,huang2023metatool} focus on laying the groundwork for datasets yet exhibit limited variety in tool usage.
To bridge this gap, \citet{qin2023toolllm} developed a more comprehensive multi-tool benchmark and proposed an advanced tool invocation method using Depth First Search-based Decision Tree (DFSDT).
Building on this, \citet{zhuang2023toolchain} employed A* search algorithm for pruning, while \citet{kim2023llm} adopted a compiler-based approach to parallelize tool invocation, both of which improved efficiency to some extent. 
Meanwhile, \citet{du2024anytool} and \citet{chen2024advancing} controlled the stability of LLM tool invocation through self-reflection and Direct Preference Optimization (DPO, \citealp{rafailov2024direct}), respectively. Despite these advances, these methods remain rooted in tree-based search paradigms, lacking a broader perspective on task planning. 
Additionally, recent works have begun exploring tool creation and integration with agents, opening new avenues for research \citep{qian2023creator,yuan2023craft,zhu2024sda,hao2024toolkengpt,schick2024toolformer,hao2024toolkengpt,ma2024event,shen2024small,yuan2024easytool}.

\paragraph{\textbf{Task-Planning for LLMs}}
Task planning capability is a crucial factor for the success of LLMs in problem-solving. Some methods attempt to decompose tasks into sub-goals and then plan for each sub-goal sequentially \citep{huang2022language,hu2023tree,lu2024chameleon,qian2024toolink,wang2024describe,shi2024learning}. HuggingGPT \citep{shen2024hugginggpt} utilizes the LLM as a controller, responsible for decomposing human-input queries into sub-tasks and ultimately generating a comprehensive response. Plan-and-Solve \citep{wang2023plan} employs a two-stage instruction prompting approach: ``Let's first devise a plan" and ``Let's carry out the plan". ProPrompt \cite{singh2023progprompt} converts natural language descriptions of problems into coding tasks. In other studies, researchers seek to interleave task decomposition and sub-task planning, advancing them dynamically \citep{gao2023pal,wu2023visual,yao2024tree,shi2025tool}. The Chain of Thought (CoT)~\citep{kojima2022large,wei2022chain} guides LLMs in reasoning about complex problems by constructing trajectories. ReAct \citep{yao2022ReAct} alternates between reasoning (the thought process) and planning (the action steps). Reflection \citep{shinn2024reflexion} builds upon ReAct by introducing a mechanism for LLMs to reflect on previous failures.

\begin{figure*}[t]
\centering
\includegraphics[width=1\textwidth]{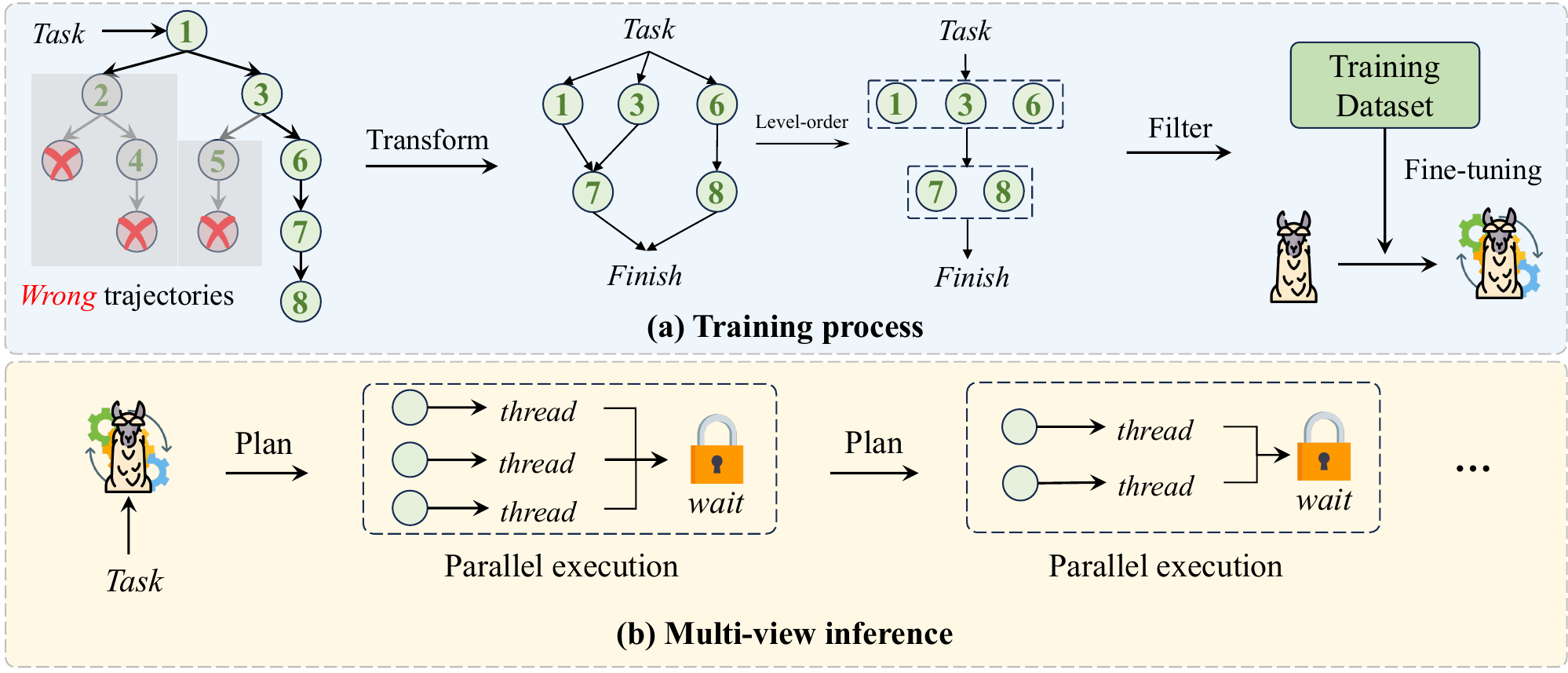}
\caption{The figure illustrates the overall DTA-Llama pipeline. (a) depicts the construction of the DTA-Tool dataset and model training; (b) shows the tool invocation inference framework based on Process/Threads.}
\label{data_construct.fig}
\end{figure*}

\section{Methodology}\label{method}
In this section, we describe: (1) the shortcoming analysis of previous methods (§ \ref{background}); (2) how the DTA-Llama be trained based on constructed parallel tool using data (§ \ref{data_cons}); (3) how to implement an efficient, Process/Threads-based parallel framework during inference (§ \ref{framework}). 

\subsection{Background}
\label{background}
As aforementioned, most recent tool invocation studies are tree-based, which are first developed by ToolLLM \cite{qin2023toolllm}.
Compared to the ReAct~\citep{yao2022ReAct} (or CoT~\citep{wei2022chain}) that invokes tools through pipelined tool interaction, the tree-based methods replace the serial tool usage with the Depth First Search-based Decision Tree (DFSDT) algorithm.
Consequentially, these methods increase the fault tolerance of LLMs and improve task planning capabilities. 
However, this comes at the cost of increased time complexity. 
This is mainly because DFSDT typically generates longer tool invocation sequences due to its backtracking mechanism, which involves multiple attempts at new nodes. 
While this improves task completion rates, it sacrifices execution efficiency (see Figure~\ref{fig:compare} for an example).


\subsection{Divide-then-Aggregate Tool Invocation}
\label{data_cons}

To address the above problems, we propose the Divide-Then-Aggregate (DTA) tool invocation paradigm. This method allows LLMs to decompose the task, generate a set of parallel tool invocations, and aggregate the results after execution. DTA improves task planning and optimizes the reasoning framework for tool invocation, enabling LLMs to invoke tools efficiently in parallel and better tackle complex tasks.


\paragraph{\textbf{Transforming the Serial Tool Using data to Parallel}}


To steer the LLMs with the capabilities of parallel tool invocation, it is critical to construct the corresponding finetuning datasets.
However, in most previous methods, e.g., vanilla CoT/ReAct methods or tree search-based algorithms like DFSDT, LLMs typically rely on invoking one tool at a time, which is not consistent with our setups.
To this end, we utilize this type of data to transform it from a serial structure into a parallel structure.


As shown in Figure \ref{data_construct.fig}, we first collect the serial successful tool innovation path from original tool searching trees.
Given the tree-like tool searching trajectories generated by the tree search-based algorithm, it is inevitable that the trajectories contain redundant or erroneous paths. Therefore, we define the node series spanning from the root to the successful leaf node as successful path \( \mathcal{P} \); we retain only the nodes in \( \mathcal{P} \), while filtering out other nodes.

Next, we utilize a powerful LLM, to identify whether any tools in \( \mathcal{P} \) can be executed in parallel. We choose GPT-4-turbo \cite{gpt4} to perform this task. If GPT-4 detects parallelizable tools, it establishes their relationships and organizes them into a Directed Acyclic Graph (DAG), represented as \( \mathcal{G} \). The feasibility of parallel execution depends on input-output dependencies and logical causal relationships.
For the nodes that cannot be parallelized, we retain their original structure\footnote{For further details on transforming to the DAG structure using GPT-4, refer to Appendix \ref{sec:prompt}. }.

Finally, we construct the tool invocation mechanism by performing the level-order traversal on \( \mathcal{G} \), enabling tools at the same level to be executed in parallel in a controlled manner and their outputs to be aggregated accordingly. This process embodies the \textit{Divide-Then-Aggregate} strategy during the data construction phase. It is worth noting that we only transform the structure of \( \mathcal{P} \) without modifying the semantic content. Therefore, the objective of our approach is not to distill or compress the data, but rather to optimize its execution structure.



\paragraph{\textbf{Data Filtering}}

To ensure data quality, we apply filtering both before and after the data transformation. 
On the one hand, we filter out the raw serial data that contain incomplete tool invocation or cause task execution failures. Because including these noisy data could negatively affect the subsequent LLMs fine-tuning.
On the other hand, we devise a rule-based filtering method to reduce some structural errors after the structural transformation. 
These rules leverage the acyclic nature of the DAG, along with its unique starting and ending points, to eliminate redundant tool invocation in cyclic invocation and situations where tool execution results cannot be aggregated.
After these steps, we obtain DTA-Tool, a high-quality, DAG-based parallel tool invocation dataset with approximately 20k entries. 
We have documented the specific filtering rules in the appendix \ref{sec:rules}.

\paragraph{\textbf{Fine-tuning}}

We fine-tune Llama-series models \cite{touvron2023llama1,touvron2023llama,dubey2024llama} using DTA-Tool. Following previous work \cite{qin2023toolllm,du2024anytool,liu2024toolace}, we employ a uniform system prompt to guide the LLMs in invoking tools based on user instructions. Details of the system prompt can be found in Appendix \ref{sec:prompt}. Our training approach has evolved from the traditional \textit{Thought-Action-Observation} framework to a streamlined \textit{Thought-Observation} framework. In this updated method, the original \textit{Action} component is integrated into \textit{Thought} as tool invocation plans. Consequently, the LLMs are trained to generate new \textit{Thoughts} by considering the user instruction alongside the history of \textit{Thoughts} and \textit{Observations}. The training loss function is defined as follows:
\begin{equation}
  \label{eq:loss}
\begin{split}
 \mathcal{L}(\theta) =& - \log \sum_{i=1}^{n} p_\theta(y^i | q,y^{[1:i-1]},o^{[1:i-1]}),
\end{split}
\end{equation}
\noindent where \( y \) represents the \textit{Thought} generated by the LLMs, \( q \) is the user instruction, and \( o \) is the \textit{Observation}. The \textit{Thought} generated in the \textit{i}-th round depends on the \textit{Thought} and \textit{Observation} from the previous round. In the final round, the focus of learning shifts from the \textit{Thought} to the final answer generated by LLMs.
Finally, we fine-tuned the LLM with the DTA-Tool, resulting in DTA-Llama, which can invoke tools in parallel.

\subsection{Process/Thread-based Inference}\label{framework} To support the modified LLMs, we developed a new inference framework, as illustrated in Figure \ref{data_construct.fig}(b). This framework redefines the \textit{Thought-Observation} cycle based on CoT/ReAct, executing tool invocation in the form of Process/Threads.

\paragraph{\textbf{Process}}
Originally, \textit{Thought} could only design an invocation strategy for a single tool, limiting its perceptual scope. In contrast, \textit{Process} enhances the LLMs' ability to \textbf{\textit{divide}} tasks and plan multiple parallelizable tool invocation strategies.
Specifically, during each round, LLMs first evaluate the task's status and progress based on the historical trajectories. Then, LLMs analyze what needs to be done in the current step and decompose the task based on the available tools. Finally, LLMs sequentially generate a series of complete tool names along with their corresponding input parameters, which can be executed in parallel.
This multi-tool approach helps broaden the perspective of LLMs, increasing the informational richness of each \textit{Thought} step.
In our framework, \textit{Process} directly integrates \textit{Action} into \textit{Thought}. After careful deliberation, the LLMs provide a formalized tool invocation plan that can be extracted using regular expressions, facilitating subsequent execution by \textit{Threads}.

\paragraph{\textbf{Threads}}
\textit{Threads} refers to the steps that faithfully execute the tool strategies presented in the \textit{Process}. In previous frameworks, \textit{Thought} provides only one tool invocation strategy. 
However, once \textit{Thought} is capable of proposing multiple tool strategies in parallel, the execution component must also support concurrent processing. To this end, we introduce \textit{Threads}.
All the tool invocation strategies provided by the \textit{Process} are distributed across multiple \textit{Thread}s, which then independently and concurrently execute each strategy.
Importantly, most real-time tool APIs inherently support a moderate degree of concurrency. Furthermore, the tool invocation plans proposed by \textit{Process} are typically lightweight. As a result, even when multiple concurrent invocations target the same API, the level of concurrency remains within a tolerable range and does not pose a risk of overloading the service.

\paragraph{\textbf{Intermediate State Lock}}
When tools are invoked using \textit{Threads}, the information processing load on the inference framework increases proportionally. The original \textit{observation} only needed to record the execution result of one tool. Now, it must systematically link multiple tools and their corresponding results in an orderly manner. Otherwise, a disorganized \textit{observation} could hinder the LLM’s subsequent decision-making.
To achieve this, we have specifically implemented a thread-oriented intermediate state lock at the end of each \textit{Threads} round.
The lock is only released once all \textit{Threads} have completed their execution and returned the results. During the complete invocation process, the intermediate state lock regularly maintains communication between \textit{Threads} and \textit{Process}. 
The execution results of \textit{Threads} are \textbf{\textit{aggregated}} and used as part of the input to interact with the LLM, initiating the next round of \textit{Process}. This cycle repeats until the task is completed.

\begin{table}[t]
\centering
\resizebox{\columnwidth}{!}{%
\begin{tabular}{@{}lr@{}}
\toprule
Statistic                                        &         \\ \midrule
\# Data scale                                    & 21,342 \\
\# Average tool invocation rounds per data                    & 2.46     \\
\# Average APIs required per data           & 3.48 \\
\% Percentage of parallel tool invocation data                   & 99.1\%    \\
\bottomrule
\end{tabular}
}
\caption{Several important characteristics of DTA-Tools are presented in the table.}
\label{tab:data}
\end{table}

\begin{table*}[t]
\centering
\resizebox{\textwidth}{!}{%
\begin{tabular}{@{}l| cc cc cc cc cc cc cc@{}}
\toprule
 & \multicolumn{2}{c}{\textbf{I1-Inst.}}
 & \multicolumn{2}{c}{\textbf{I1-Tool}}
 & \multicolumn{2}{c}{\textbf{I1-Cat.}} 
 & \multicolumn{2}{c}{\textbf{I2-Inst.}} 
 & \multicolumn{2}{c}{\textbf{I2-Cat.}} 
 & \multicolumn{2}{c}{\textbf{I3-Inst.}} 
 & \multicolumn{2}{c}{\textbf{Average}}
 \\ \cmidrule(l){2-15} 
\multirow{-2}{*}{\textbf{Method}} 
& SoPR & \multicolumn{1}{c|}{SoWR}
& SoPR & \multicolumn{1}{c|}{SoWR}
& SoPR & \multicolumn{1}{c|}{SoWR}
& SoPR & \multicolumn{1}{c|}{SoWR} 
& SoPR & \multicolumn{1}{c|}{SoWR}
& SoPR & \multicolumn{1}{c|}{SoWR} 
& SoPR & SoWR \\
\hline
\rowcolor{Gainsboro} \multicolumn{15}{c}{\textit{GPT-series}} \\
\midrule 
\specialrule{0em}{1pt}{1pt}

GPT-3.5 (ReAct)
& 53.0 & \multicolumn{1}{c|}{--} 
& 53.0 & \multicolumn{1}{c|}{--}
& 51.2 & \multicolumn{1}{c|}{--}
& 37.6 & \multicolumn{1}{c|}{--} 
& 43.9 & \multicolumn{1}{c|}{--}
& 48.6 & \multicolumn{1}{c|}{--} 
& 47.9 & - \\

GPT-3.5 (DFSDT) 
& 63.8 & \multicolumn{1}{c|}{58.9}
& 73.9 & \multicolumn{1}{c|}{65.8}
& 65.8 & \multicolumn{1}{c|}{60.1}
& 57.1 & \multicolumn{1}{c|}{72.6} 
& 69.8 & \multicolumn{1}{c|}{68.5} 
& 69.9 & \multicolumn{1}{c|}{67.2} 
& 66.7 & 65.5 \\

GPT-3.5 (Parallel) 
& 64.6 & \multicolumn{1}{c|}{48.5}
& 65.0 & \multicolumn{1}{c|}{55.7} 
& 69.0 & \multicolumn{1}{c|}{54.2}
& 54.9 & \multicolumn{1}{c|}{55.7}
& 61.4 & \multicolumn{1}{c|}{53.2}
& 56.6 & \multicolumn{1}{c|}{50.8} 
& 61.9 & 53.0 \\

GPT-4 (ReAct) 
& 54.4 & \multicolumn{1}{c|}{53.4}
& 44.1 & \multicolumn{1}{c|}{60.1} 
& 48.8 & \multicolumn{1}{c|}{52.9} 
& 50.6 & \multicolumn{1}{c|}{69.8} 
& 48.9 & \multicolumn{1}{c|}{62.1}
& 42.6 & \multicolumn{1}{c|}{54.1}
& 48.2 & 58.7 \\

GPT-4 (DFSDT) 
& \best{69.0} & \multicolumn{1}{c|}{57.1} 
& \best{69.6} & \multicolumn{1}{c|}{\best{66.5}}
& 68.1 & \multicolumn{1}{c|}{61.4} 
& 70.8 & \multicolumn{1}{c|}{73.6} 
& 68.0 & \multicolumn{1}{c|}{62.9}
& \best{76.0 }& \multicolumn{1}{c|}{63.9} 
&  \best{70.3} & 64.2 \\

GPT-4 (Parallel) 
& 62.9 & \multicolumn{1}{c|}{\best{66.3}} 
& 67.4 & \multicolumn{1}{c|}{61.4} 
& \best{70.9} & \multicolumn{1}{c|}{\best{62.7}}
&  \best{73.4} & \multicolumn{1}{c|}{\best{85.8}}
& \best{70.8} & \multicolumn{1}{c|}{\best{77.4}}
& 69.7 & \multicolumn{1}{c|}{\best{70.5}} 
& 69.2 &  \best{70.7} \\
\hline

\rowcolor{Gainsboro} \multicolumn{15}{c}{\textit{Open-source}} \\
\midrule
\specialrule{0em}{1pt}{1pt}

ToolLLaMA (ReAct)
& 42.7 & \multicolumn{1}{c|}{36.2}
& 35.4 & \multicolumn{1}{c|}{36.1}
& 38.6 & \multicolumn{1}{c|}{34.6} 
& 39.9 & \multicolumn{1}{c|}{49.1}
& 40.9 & \multicolumn{1}{c|}{38.7} 
& 29.8 & \multicolumn{1}{c|}{41.0}
& 37.9 & 39.3 \\

\ \ {----\small{ToolLLaMA$\dag$ (ReAct)}}
& 26.7 & \multicolumn{1}{c|}{22.1} 
& 25.0 & \multicolumn{1}{c|}{27.2}
& 31.7 & \multicolumn{1}{c|}{29.4}
& 23.1 & \multicolumn{1}{c|}{32.1} 
& 24.5 & \multicolumn{1}{c|}{28.2}
& 20.5 & \multicolumn{1}{c|}{24.6} 
& 25.3 & 27.3 \\

ToolLLaMA (DFSDT)
& 56.6 & \multicolumn{1}{c|}{39.9} 
& 55.5 & \multicolumn{1}{c|}{46.8} 
& 56.5 & \multicolumn{1}{c|}{41.8}
& 49.7 & \multicolumn{1}{c|}{53.8}
& 53.4 & \multicolumn{1}{c|}{49.2} 
& 53.6 & \multicolumn{1}{c|}{50.8}
& 54.2 & 47.1 \\

\ \ {----\small{ToolLLaMA$\dag$ (DFSDT)}}
& 41.8 & \multicolumn{1}{c|}{35.6} 
& 39.9 & \multicolumn{1}{c|}{37.3}
& 44.9 & \multicolumn{1}{c|}{39.9}
& 36.0 & \multicolumn{1}{c|}{47.2} 
& 39.1 & \multicolumn{1}{c|}{39.5} 
& 33.3 & \multicolumn{1}{c|}{26.2} 
& 39.2 & 37.6 \\

LLMCompiler
& 39.2 & \multicolumn{1}{c|}{35.6} 
& 35.1 & \multicolumn{1}{c|}{36.0}
& 39.8 & \multicolumn{1}{c|}{35.3}
& 37.5 & \multicolumn{1}{c|}{45.6} 
& 38.4 & \multicolumn{1}{c|}{38.1} 
& 27.0 & \multicolumn{1}{c|}{36.5} 
& 36.2 & 37.9 \\

Qwen2.5 (Parallel) 
& \high{65.7} & \multicolumn{1}{c|}{\high{54.0}} 
& 58.8 & \multicolumn{1}{c|}{51.0}
& 63.5 & \multicolumn{1}{c|}{52.4}
& 60.2 & \multicolumn{1}{c|}{55.6} 
& 61.3 & \multicolumn{1}{c|}{61.3} 
& \high{68.3} & \multicolumn{1}{c|}{57.6} 
& 63.0 & 55.3 \\

Ours 
& 63.5 & \multicolumn{1}{c|}{52.1}
& \high{64.2} & \multicolumn{1}{c|}{\high{53.2}}
& \high{67.2} & \multicolumn{1}{c|}{\high{54.2}}
& \high{62.1} & \multicolumn{1}{c|}{\high{70.8} }
&  \high{71.9} & \multicolumn{1}{c|}{\high{65.3}}
& 67.5  & \multicolumn{1}{c|}{\high{59.0}}
& \high{66.1} & \high{59.1} \\
\bottomrule
\end{tabular}
}
\caption{A comparison different baselines and method on StableToolBench. Considering that real-world APIs are time-sensitive, the results of baselines presented in the table are reproduced during the period \texttt{from September to October 2024} using their official implementation. We highlight the best performance of GPT-series models and open-source models with the \besttext{red} and \hightext{blue}, respectively.}
\label{tab:main}
\end{table*}

\section{Experimental Setup}

\paragraph{\textbf{Dataset}}


We use StableToolBench \cite{guo2024stabletoolbench} for evaluation.
All test cases in StableToolBench are actually derived from the test portion of ToolBench~\cite{touvron2023llama}.
Concretely, ToolBench is divided into six evaluation subsets based on tool categories and scenarios. The tool categories are as follows: \textit{Inst.} denotes unseen instructions for the same set of tools in the training data, \textit{Tool} denotes unseen tools within the same (seen) category as those in the training data, and \textit{Cat.} denotes unseen tools from a different (unseen) category. The scenarios are: \textit{I1} for single-tool instructions, \textit{I2} for intra-category multi-tool instructions, and \textit{I3} for intra-collection multi-tool instructions. 
The difficulty level of the task escalates progressively from \textit{I1-Inst.} to \textit{I3-Inst.}.
Compared to the original ToolBench, StableToolBench introduces an extra caching system and an API simulator to mitigate the instability issues associated with real-time APIs.

As for the DTA-Llama training, we adopt the training portion of ToolBench and transform it into DTA-Tool style using GPT-4-turbo \cite{gpt4} through the method described in Sec.\ref{data_cons}. ToolBench provides a corresponding API set for each data point, enabling us to focus on tool learning without having to pay attention to tool retrieval.
A detailed overview of DTA-Tool is presented in Table \ref{tab:data}, with an instance provided in Appendix \ref{sec:instance}. 


\paragraph{\textbf{Baselines}}
We use both GPT-series and other open-source LLMs as our baselines. For the GPT-series models, we use OpenAI's GPT-3.5-turbo and GPT-4-turbo, leveraging their function calling capabilities\footnote{While the exact mechanisms remain unclear, OpenAI has enabled parallel tool invocation in these models.}. And we include Parallel as an additional baseline paradigm alongside ReAct and DFSDT. 

For open-source models, we fine-tune ToolLLaMA~\citep{qin2023toolllm} from Llama2-7B on ToolBench and compare it using the ReAct and DFSDT methods. 
LLMCompiler~\cite{kim2023llm} is a non-training-based parallel tool invocation method that relies on system design and prompt engineering. We also use it as a baseline for comparison with Llama2-7B.
Additionally, recent open-source LLMs, such as Qwen2.5 \cite{qwen2.5}, have demonstrated strong capabilities, including function calls and parallel tool invocation. Consequently, we include Qwen2.5-7B-Instruct in our baseline for comparison. To ensure experimental consistency, our method is fine-tuned on Llama2-7B for a fair comparison.

\paragraph{\textbf{Evaluation Metrics}}

StableToolBench introduces two key metrics to assess the tool learning capabilities of LLMs: Solvable Pass Rate (SoPR) and Solvable Win Rate (SoWR). SoPR measures the success rate of LLMs in solving tasks, while SoWR compares the quality of results against the GPT-3.5 (ReAct) baseline.
All the results are averaged from three tests to minimize variance.

Besides the SoPR and SoWR, we expanded the StableToolBench evaluation to further investigate the efficiency of LLMs.
We assess the computational cost of LLMs from two dimensions: token consumption and inference time, to better analyze their efficiency in task-solving.

\paragraph{\textbf{Implementation Details}}

We use the Llama-series models \cite{touvron2023llama1,touvron2023llama,dubey2024llama} as our backbone, and fine-tune them for our task. Since Llama2 has a context length limit of 4096 tokens, shorter contexts may not be sufficient for effective tool invocation. To overcome this limitation, we followed the approach in ToolLLaMA~\cite{qin2023toolllm} and applied position interpolation \cite{chen2023extending} to extend the context length to 8192 tokens. More details about our training process can be found in Appendix \ref{sec:train}.

\begin{figure*}[t]
\centering
\includegraphics[width=1\textwidth]{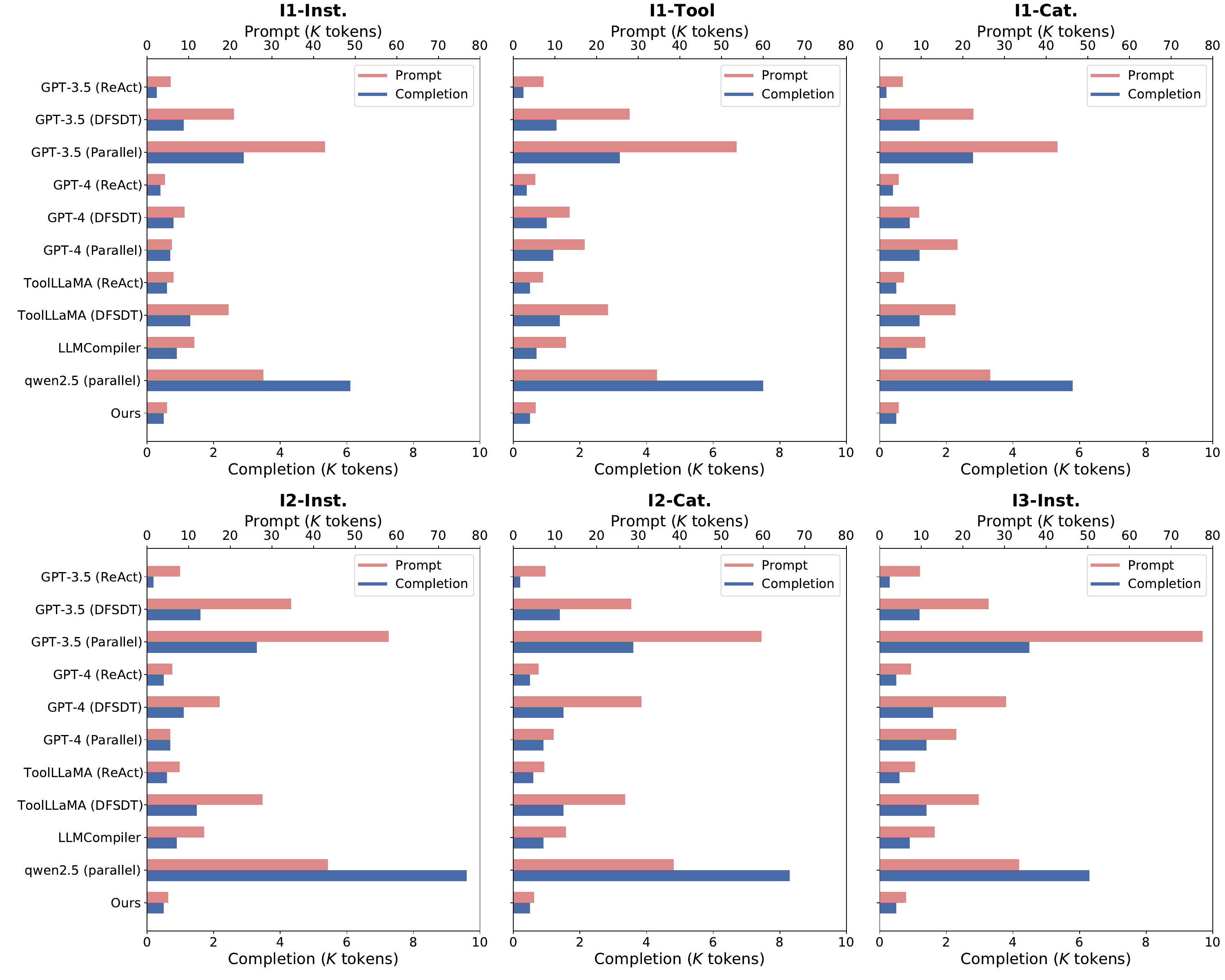}
\caption{A comparison of all methods in terms of token consumption. The figure has two horizontal axes, representing \textit{Prompt} and \textit{Completion}, both measured in thousand tokens ($\mathit{K}$ tokens)).}
\label{fig:token}
\vspace{-0.5em}
\end{figure*}

\section{Experiments}
In this section, we first evaluate the performance of our method in tool learning tasks through extensive experiments in \S~\ref{main_exp}. Next, we analyze its computational costs compared to baselines in \S~\ref{coump_exp}, and extend our method to different models to evaluate its generalizability in \S~\ref{generalizability}. We showcase the practical workflow of DTA-Llama through case examples in Appendix~\ref{sec:case}.

\subsection{Main Experiments}\label{main_exp}

\paragraph{\textbf{SoPR}}

As shown in Table \ref{tab:main}, our method surpasses all open-source baselines~\footnote{
To study the effect of training data, for each sample in the DTA-Tool, we collect its raw data in ToolBench and merge them together (denoted as DTA-Tool*); then,  we re-train Llama2-7B on DTA-Tool* using the ReAct and DFSDT methods, denoting this version with $\dag$.
However, the performance of this version is even inferior to the original, suggesting that data filtering is not the primary reason for the contribution.}. 
While GPT series models, particularly GPT-4, show superior performance when compared to earlier open-source models, our approach not only surpasses GPT-3.5 but also competes with GPT-4. These results suggest that our method has a notable impact on enhancing the tool invocation capabilities of LLMs.

Moreover, under the same GPT model conditions, DFSDT and Parallel demonstrate similar performances, suggesting that their parallel function call strategy does not notably enhance the tool invocation capabilities of GPT-based LLMs. In contrast, Qwen2.5 (Parallel) exhibited a clear improvement over DFSDT in open-source LLMs, with our method further advancing this performance. This demonstrates that our parallel mechanism is more effective than other parallel approaches, providing a greater boost to tool invocation capabilities.


\paragraph{\textbf{SoWR}}

The Solvable Pass Rate evaluates the quality of results against the GPT-3.5 (ReAct) baseline. As shown in Table~\ref{tab:main}, most open-source models achieve a SoWR score below 50, indicating their responses are of lower quality compared to GPT-3.5 (ReAct). In contrast, our method outperforms all open-source models, achieving a SoWR score of 70.8 on the I2-Inst dataset (27.32\% relative improvement). 
Compared to GPT-3.5 DFSDT \& Parallel, our method achieves nearly equivalent performance based solely on Llama2-7B, and even surpasses it in certain subsets.

These results, combined with SoPR, demonstrate that our method significantly enhances the ability of LLMs' tool utilization. We attribute this improvement to our specialized parallel tool invocation mechanism. This mechanism expands the decision-making scope of LLMs within one round, reducing the overall trajectories while enabling better decision-making. We provide example of the actual decision-making processes of different methods in the Appendix~\ref{sec:case} to illustrate this.



\subsection{Computational Cost}\label{coump_exp}

\paragraph{\textbf{Token Consumption}}
We count tokens for both \textit{Completion} (output tokens generated by LLMs) and \textit{Prompt} (input tokens provided to LLMs). Typically, \textit{Completion} is more costly than \textit{Prompt}.  
Figure \ref{fig:token} shows the token consumption for all methods. 
For each subset of StableToolBench, we calculate the average token count across all cases to assess the overall performance of the LLMs.
The results in Figure \ref{fig:token} show that our approach is highly competitive among open-source models. It significantly outperforms DFSDT while costing less than ReAct. Compared to GPT-based methods, our approach is particularly cost-effective, consuming fewer tokens. Given that DTA-Llama is a 7-billion-parameter model, its actual deployment costs are even lower.
In addition, we also provide statistics on the maximum token consumption for each subset in Appendix~\ref{sec:supply_exp}, to assess the performance of the LLMs in handling complex scenarios.


Furthermore, we analyze the inference steps of each baseline by calculating the average number of steps across all subsets.
As demonstratd in Table \ref{tab:step}, our method maintains its strong performance in terms of token consumption, requiring the fewest inference steps. 
In contrast, both the GPT series and Qwen2.5 models exhibit higher inference steps with their Parallel methods, which may be attributed to their limited task-planning capabilities.
The LLMCompiler, a non-training approach, faces a bottleneck in processing complex tasks with 7-billion-parameter models. When considered alongside the results presented in Table \ref{tab:main}, its performance falls short of expectations.

\begin{table}[tp]
\centering
\resizebox{\columnwidth}{!}{%
\setlength{\tabcolsep}{2pt} 
\begin{tabular}{@{}
                p{3.7cm} |
                >{\centering\arraybackslash}p{1.3cm} 
                >{\centering\arraybackslash}p{1.3cm} 
                >{\centering\arraybackslash}p{1.3cm} 
                >{\centering\arraybackslash}p{1.3cm} 
                >{\centering\arraybackslash}p{1.3cm} 
                >{\centering\arraybackslash}p{1.3cm} 
                @{}}
\toprule
\textbf{Method} &
  \textbf{I1-Inst.} &
  \textbf{I1-Tool} &
  \textbf{I1-Cat.} &
  \textbf{I2-Inst.} &
  \textbf{I2-Cat.} &
  \textbf{I3-Inst.} \\
\midrule 
\rowcolor{Gainsboro} \multicolumn{7}{c}{GPT-series} \\
\midrule 
GPT-3.5 (ReAct)      & 4.28  & 4.75  & 4.48  & 5.16  & 5.05  & 5.31 \\
GPT-3.5 (DFSDT)    & 11.60 & 13.36 & 11.77 & 16.60 & 14.06 & 12.54 \\
GPT-3.5 (parallel) & 25.33 & 28.06 & 26.12 & 31.79 & 31.04 & 38.10 \\
GPT-4 (ReAct)        & 3.27  & 3.64  & 3.87  & 4.04  & 4.19  & 4.23  \\
GPT-4 (DFSDT)      & 5.90  & 8.09  & 6.67  & 9.97 & 18.13 & 14.05 \\
GPT-4 (parallel)   & 4.66 & 9.18 & 12.90  & 3.63 & 5.98 & 10.38  \\
\midrule 
\rowcolor{Gainsboro} \multicolumn{7}{c}{Open-source} \\
\midrule 
ToolLLaMA (ReAct)    & 3.42  & 3.47  &   3.50  & 3.67  & 3.63  & 3.64  \\
ToolLLaMA (DFSDT)  & 8.09  & 8.51  & 8.10 & 10.20  & 9.93  & 9.23  \\
LLMCompiler &  5.48  & 5.56  & 6.07  & 5.36  & 5.68  & 5.62  \\
Qwen2.5 (Parallel) &  9.07  & 9.47  & 12.01  & 14.58  & 14.56  & 12.38  \\
\textbf{Ours} & \high{2.41}  & \high{2.41}  & \high{2.51}  & \high{2.32}  & \high{2.34}  & \high{2.48}  \\ \bottomrule
\end{tabular}
}
\caption{A comparison of the number of inference steps across different methods.}
\label{tab:step}
\end{table}

\begin{table}[]
\centering
\resizebox{\columnwidth}{!}{%
\begin{tabular}{l | ccccccc}
\toprule
\textbf{Dataset} & ReAct & DFSDT & LLMCompiler & Qwen2.5 & Ours \\
\toprule
\rowcolor{Gainsboro}  \multicolumn{6}{c}{Inference latency (s)} \\
\midrule
\textbf{I1-Inst.}  &  34  & 76$_{\uparrow 124\%}$   & 58$_{\uparrow 71\%}$   & 104$_{\uparrow 205\%}$ & \high{29$_\mathbf{\downarrow 15\%}$}  \\
\textbf{I1-Tool}  &  40  &  103$_{\uparrow 158\%}$   &  67$_{\uparrow 68\%}$   & 111$_{\uparrow 177\%}$ & \high{33$_\mathbf{\downarrow 20\%}$} \\
\textbf{I1-Cat.}    &  35   & 90$_{\uparrow 157\%}$  & 61$_{\uparrow 74\%}$    & 84$_{\uparrow 140\%}$ & \high{28$_\mathbf{\downarrow 20\%}$} \\
\textbf{I2-Inst.}  &  39   &  110$_{\uparrow 182\%}$   &  83$_{\uparrow 113\%}$   & 153$_{\uparrow 292\%}$  & \high{29$_\mathbf{\downarrow 26\%}$} \\
\textbf{I2-Cat.}  &   38  & 124$_{\uparrow 226\%}$    & 71$_{\uparrow 87\%}$   & 130$_{\uparrow 241\%}$ & \high{30$_\mathbf{\downarrow 21\%}$} \\
\textbf{I3-Inst.}  &  46  & 120$_{\uparrow 161\%}$    & 69$_{\uparrow 50\%}$   & 135$_{\uparrow 193\%}$ & \high{40$_\mathbf{\downarrow 13\%}$} \\
\textbf{Avg.}  &  39  & 104$_{\uparrow 167\%}$ & 68$_{\uparrow 74\%}$ & 119$_{\uparrow 204\%}$  & \high{31$_{\downarrow 21\%}$} \\
\bottomrule
\rowcolor{Gainsboro}  \multicolumn{6}{c}{{ Speed up (rate)}} \\
\midrule 
\textbf{I1-Inst.}  &  1.00  & $\times$0.45  &  $\times$0.59  & $\times$0.33 &  \high{$\times$1.18} \\
\textbf{I1-Tool} &   1.00  & $\times$0.39  &  $\times$0.60  & $\times$0.36 &   \high{$\times$1.20} \\
\textbf{I1-Cat.}   & 1.00    & $\times$0.39  &  $\times$0.57  & $\times$0.42 &  \high{$\times$1.20} \\
\textbf{I2-Inst.} &   1.00  & $\times$0.31  &  $\times$0.47  & $\times$0.26 &  \high{ $\times$1.27 }\\
\textbf{I2-Cat.} &   1.00  &  $\times$0.35  &  $\times$0.54  & $\times$0.29 &  \high{$\times$1.35} \\
\textbf{I3-Inst.} &   1.00  & $\times$0.38  &  $\times$0.67  & $\times$0.34 &  \high{$\times$1.15}  \\
\textbf{Avg.} &   1.00  & $\times$0.37  &  $\times$0.57  & $\times$0.33 &  \high{$\times$1.22} \\
\bottomrule
\end{tabular}
}
\caption{A comparison of the inference times of LLMs. Inference latency represents the average inference time across all cases in different subsets; Speed up indicates the factor of the inference speed being improved relative to ToolLLaMA (ReAct).}
\label{tab:time}
\end{table}

\paragraph{\textbf{Inference Time}}
Inference time is another key metric for assessing the computational cost of LLMs, which directly influences the deployment overhead of the service. 
Considering the intangibility of the GPT-series models and the impact of network latency on service requests, our inference time experiments focus solely on open-source LLMs. 
Similar to the setting of token consumption comparisons, we record the inference times for all cases within each subset and calculated the average.
As shown in Table \ref{tab:time}, our method demonstrates a clear advantage in inference time.
Furthermore, combining the results from Table \ref{tab:main}, we can conclude that our method delivers better performance at lower computational costs.

\begin{figure}[t]
\centering
\includegraphics[width=\columnwidth]{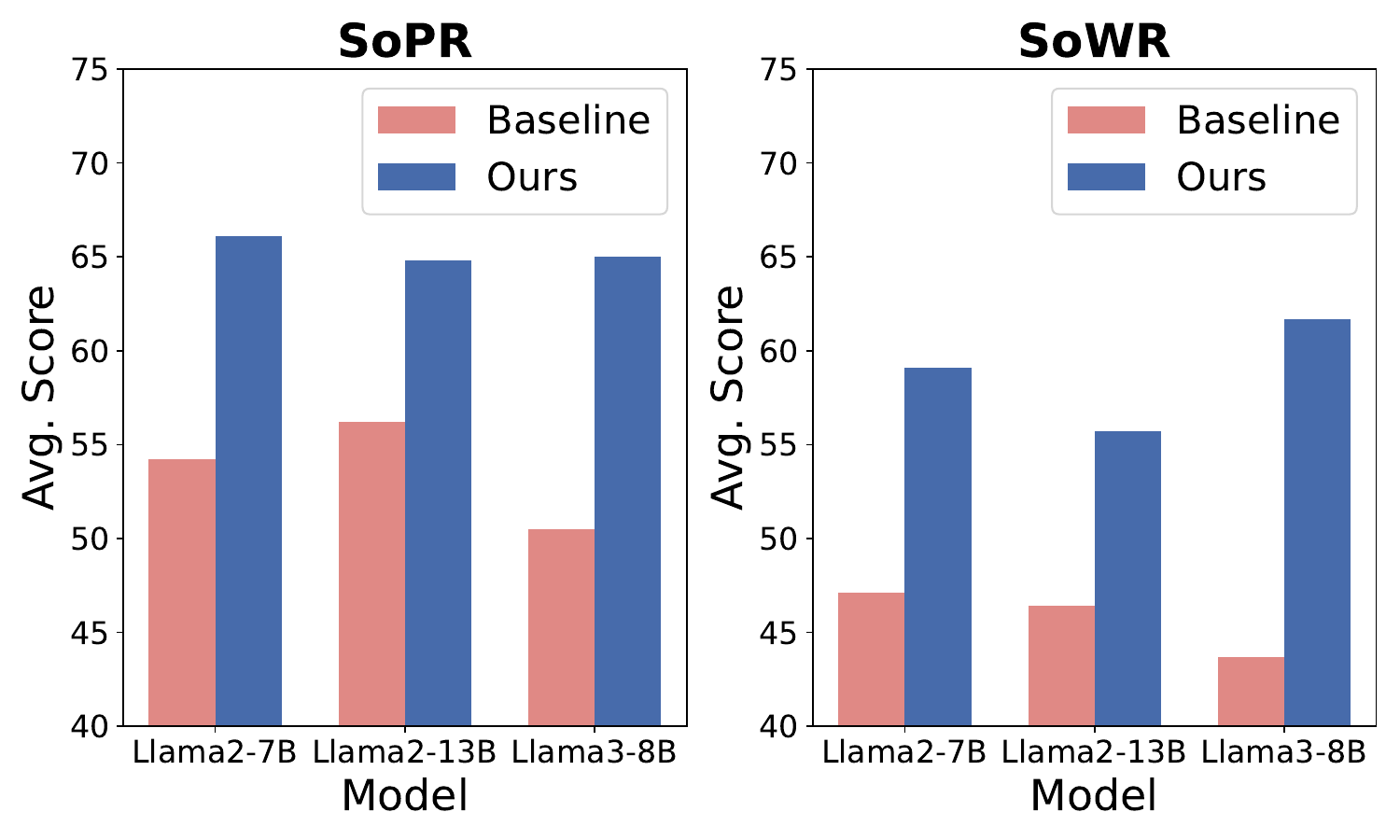}
\caption{A comparison of different LLMs after fine-tuning with our method and the baseline method.}
\label{fig:general}
\end{figure}

\subsection{Generalizability}\label{generalizability}


To validate whether our method can achieve similar improvement across a wide range of LLMs besides Llama2-7B, we select ToolLLaMA (DFSDT) as the baseline and conduct fine-tuning and testing on Llama2-13B and Llama3-8B. 
The average experimental results of six benchmark subsets are presented in Figure \ref{fig:general}.
The Llama2-7B results are derived from the \textit{Averages} of ToolLLaMA (DFSDT) and DTA-Llama in Table \ref{tab:main}. 
The results show that our method significantly outperforms the baselines across across all scales of LLMs, especially in the SoWR metric for Llama3-8B, where the improvement exceeds 40\%.
More detailed experimental results and analysis are provided in Appendix \ref{sec:supply_exp}.

\section{Conclusion}

In this paper, we introduce DTA-Llama, a novel tool learning approach based on the parallel invocation of tools through the iteratively Divide-Then-Aggregate paradigm. We construct the training data by transforming sequential data into a parallel DAG structure and use this data to train the model. Subsequently, we integrate a Process/Threads-based inference framework to enable LLMs to perform tool invocation in parallel. Extensive experimental results demonstrate that, compared to existing methods, DTA-Llama not only significantly improves performance but also substantially enhances the efficiency of tool learning in LLMs.

\section*{Limitations}




This paper aims to advance the research of tool learning in LLMs, particularly in both industry and academia. However, due to limitations in human resources, computational power, and the current research conditions, there are certain constraints, as outlined below:

First, due to resource constraints, we were unable to conduct additional experiments on larger models to further validate the effectiveness of DTA-Llama. With the publication of this paper, we hope that more researchers in the field will attempt to build upon and extend our work.

Second, although our method shows improvements over existing tool learning approaches, LLMs still struggle to reliably and consistently address complex real-world problems through tool invocation. We hope to attract more researchers to the study of tool learning, as this area urgently requires more attention and resources.

\bibliography{custom}

\clearpage

\appendix

\section{Prompt}
\label{sec:prompt}

\begin{figure*}[tp]
\centering
\includegraphics[width=1\textwidth]{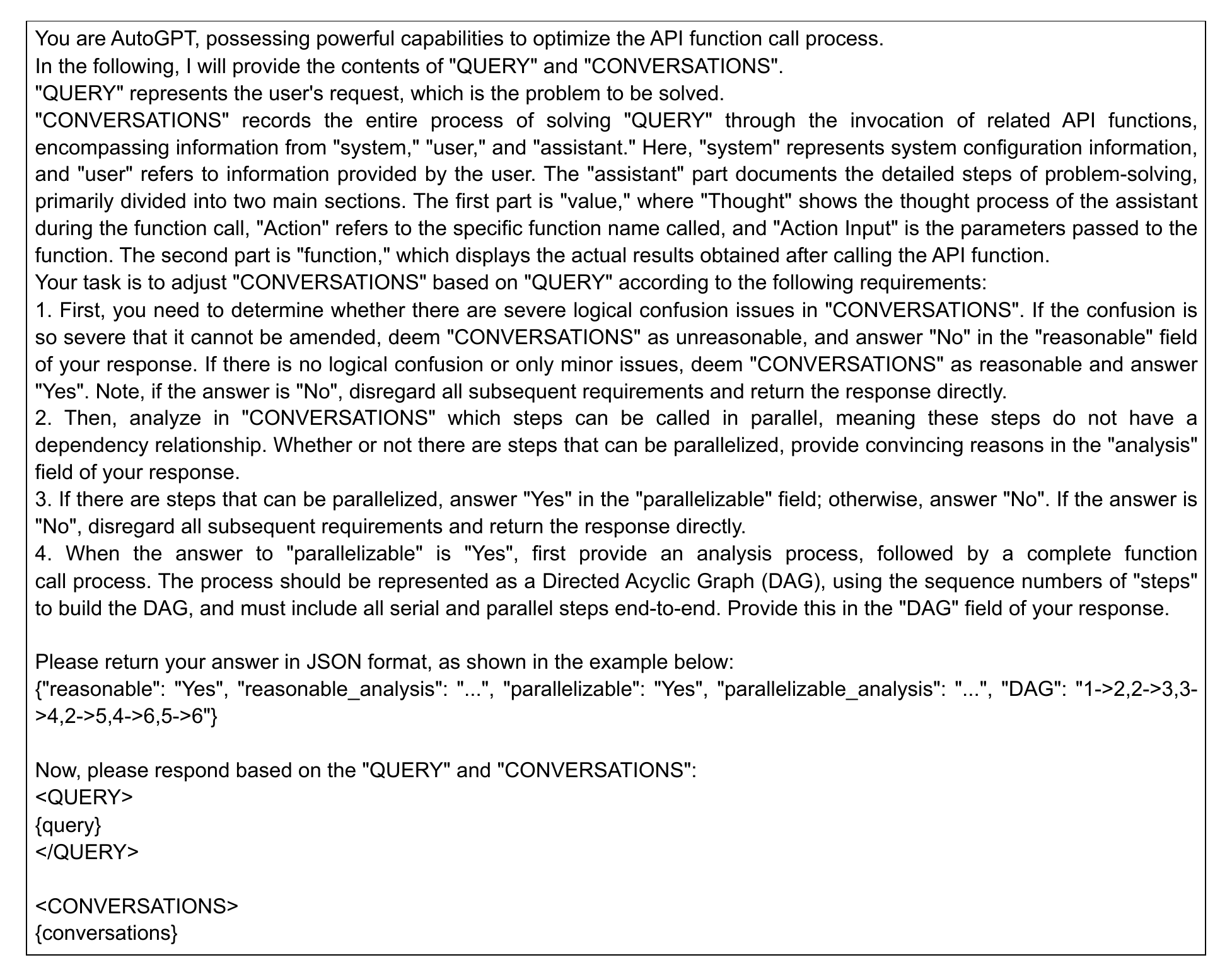}
\caption{The prompt used for data transformation: in this context, \textit{\{query\}} represents the instruction from the user, while \textit{\{conversations\}} refers to the original conversation content in the ToolBench data.}
\label{data_prompt.fig}
\end{figure*}

\paragraph{\textbf{Data Transformation Prompt}}
Figure \ref{data_prompt.fig} shows the prompts used during the data transformation process. We use this prompt with GPT-4-turbo to convert the serial data into a DAG structure. The prompt first evaluates the reasonableness of the conversation, discarding any data that doesn’t meet the criteria. It then assesses whether the steps can be processed in parallel, generating a planning path within the DAG structure. The evaluations of reasonableness and parallelism follow distinct analytical processes, reflecting a form of internalized CoT prompt engineering. By applying these analyses to DAG generation, we achieve more accurate results.

\begin{figure*}[tp]
\centering
\includegraphics[width=1\textwidth]{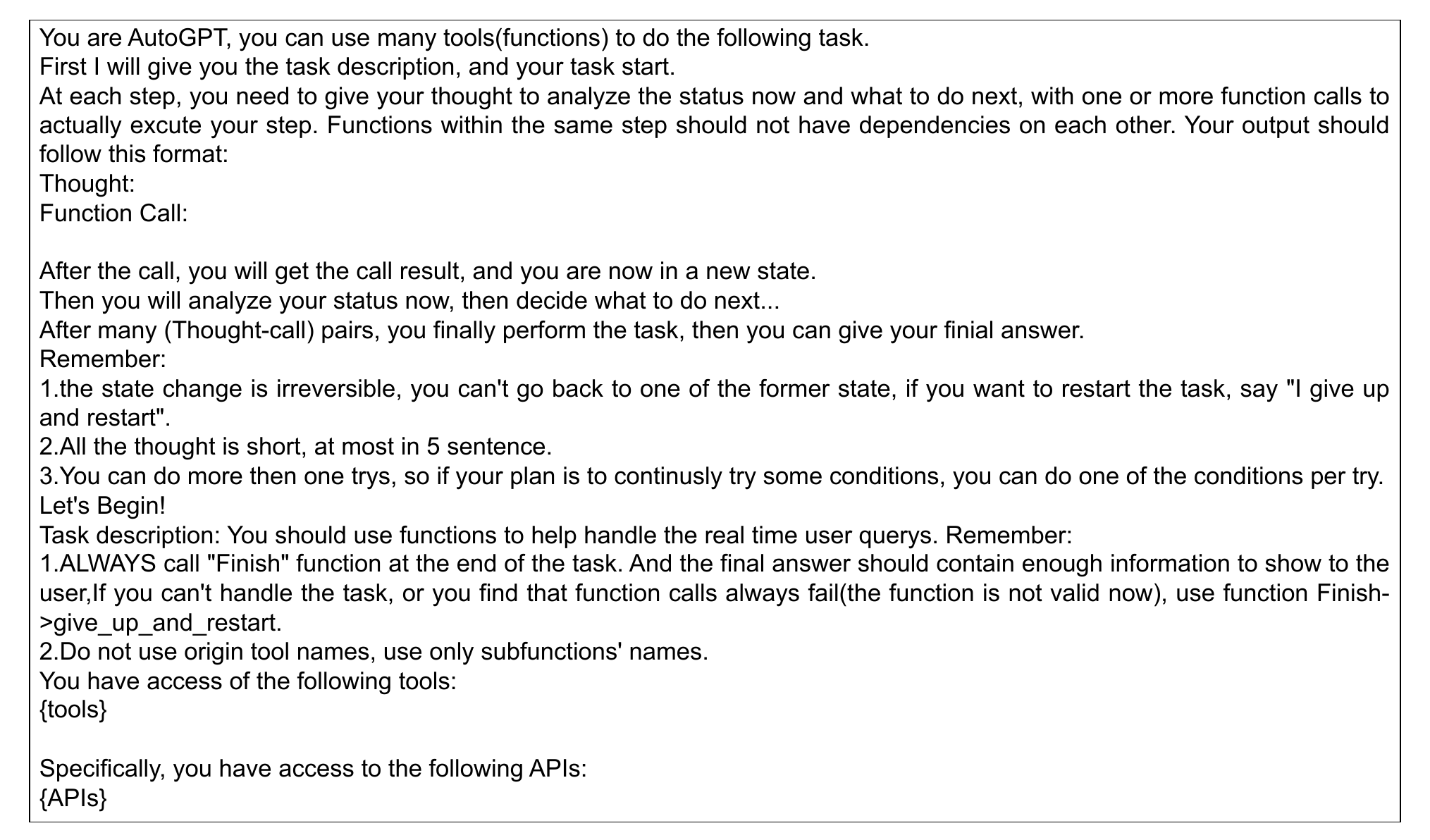}
\caption{A unified system prompt is used during both training and inference. It must be input into the LLMs at the beginning of the conversation.}
\label{infer_prompt.pdf}
\end{figure*}

\paragraph{\textbf{System Prompt}}
Figure \ref{infer_prompt.pdf} illustrates the system prompt used during both training and inference. Each user instruction is paired with a list of tool candidates, including both relevant and unrelated. The \textit{tools} section contains the names and descriptions of all tools in the candidate set. Additionally, each tool is equipped with a set of APIs designed to handle various types of tasks. The \textit{API} section is represented as a JSON list, providing detailed information about the names, descriptions, and parameters of the APIs associated with each tool.

\section{Data Filtering Rules}\label{sec:rules}
\begin{figure*}[tp]
\centering
\includegraphics[width=1\textwidth]{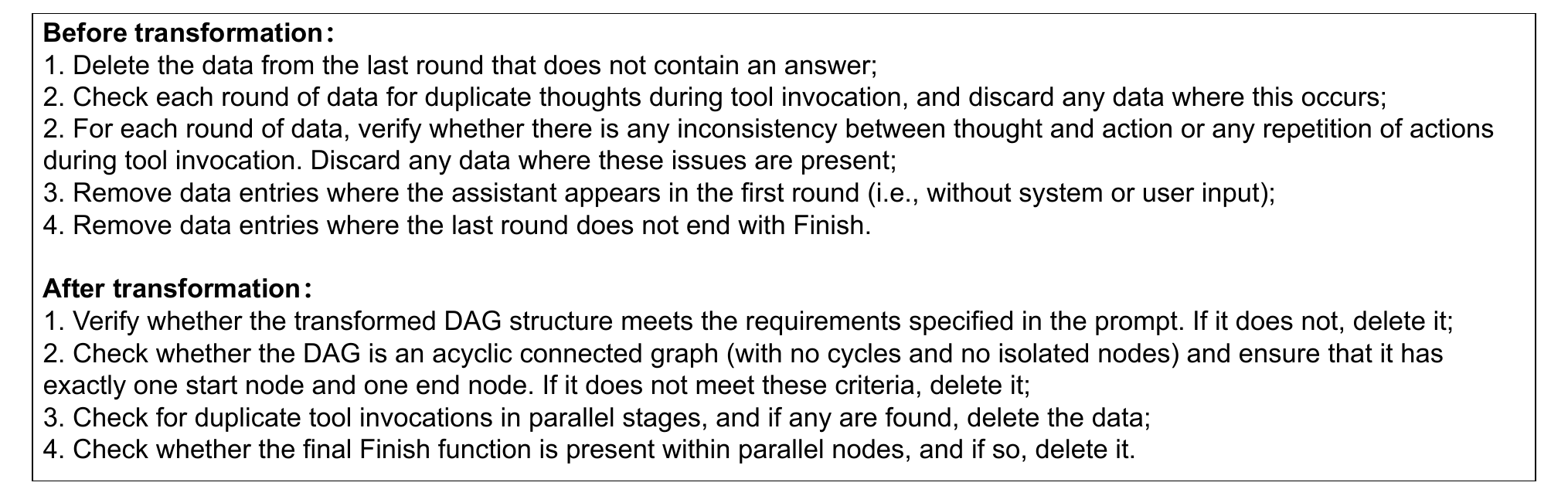}
\caption{The image illustrates the detailed data filtering rules.}
\label{rules.pdf}
\end{figure*}

The rules for data filtering are divided into before and after structural transformation. The specific rules are listed in text form in Figure \ref{rules.pdf}. These rules can be easily implemented in code to filter the data.

\section{DTA-Tool Instance}\label{sec:instance}

\begin{figure*}[tp]
\centering
\includegraphics[width=1\textwidth]{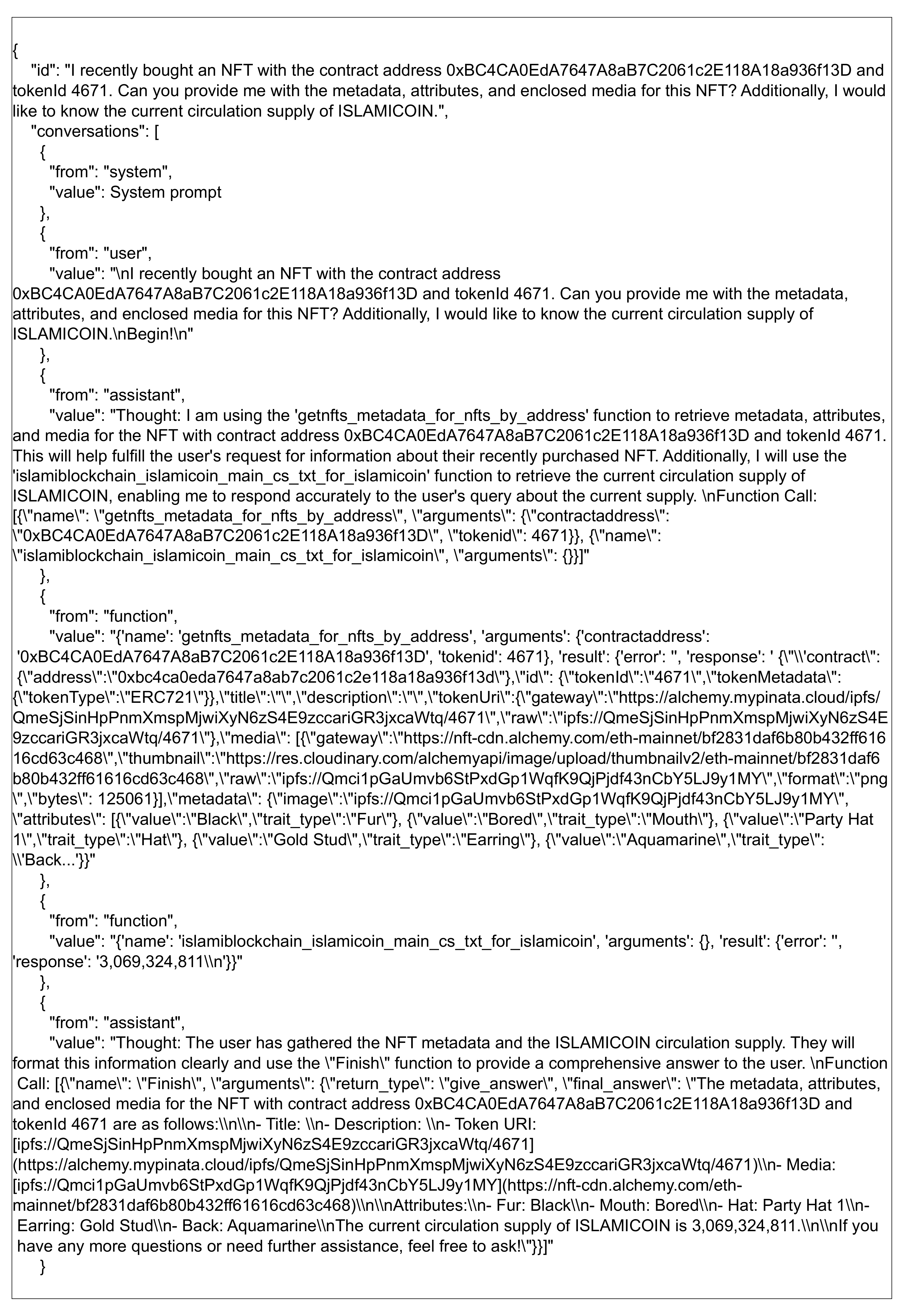}
\caption{An instance of a data entry in DTA-Tool.}
\label{instance.pdf}
\end{figure*}

Figure \ref{instance.pdf} illustrates a data entry in DTA-Tool, stored in JSON format. The outer layer consists of two keys: ``id'' and ``conversations''. The ``id'' represents the user's instruction, while ``conversations'' details the task execution process by the LLMs. Within ``conversations'', there are four roles: ``system'', ``user'', ``assistant'', and ``function''. The ``system'' role, which represents the system prompt, is introduced in Appendix \ref{sec:prompt} and is omitted here; ``user'' is the same as ``id'' and reflects the user's instruction; ``assistant'' represents the LLMs' reasoning process (marked by \textit{Thought}) and provides a specific tool invocation plan (marked by \textit{Function Call}); ``function'' contains the result of the tool execution. When tool invocation are parallel, the results are concatenated sequentially.

\begin{figure*}[t]
\centering
    \includegraphics[width=1\textwidth]{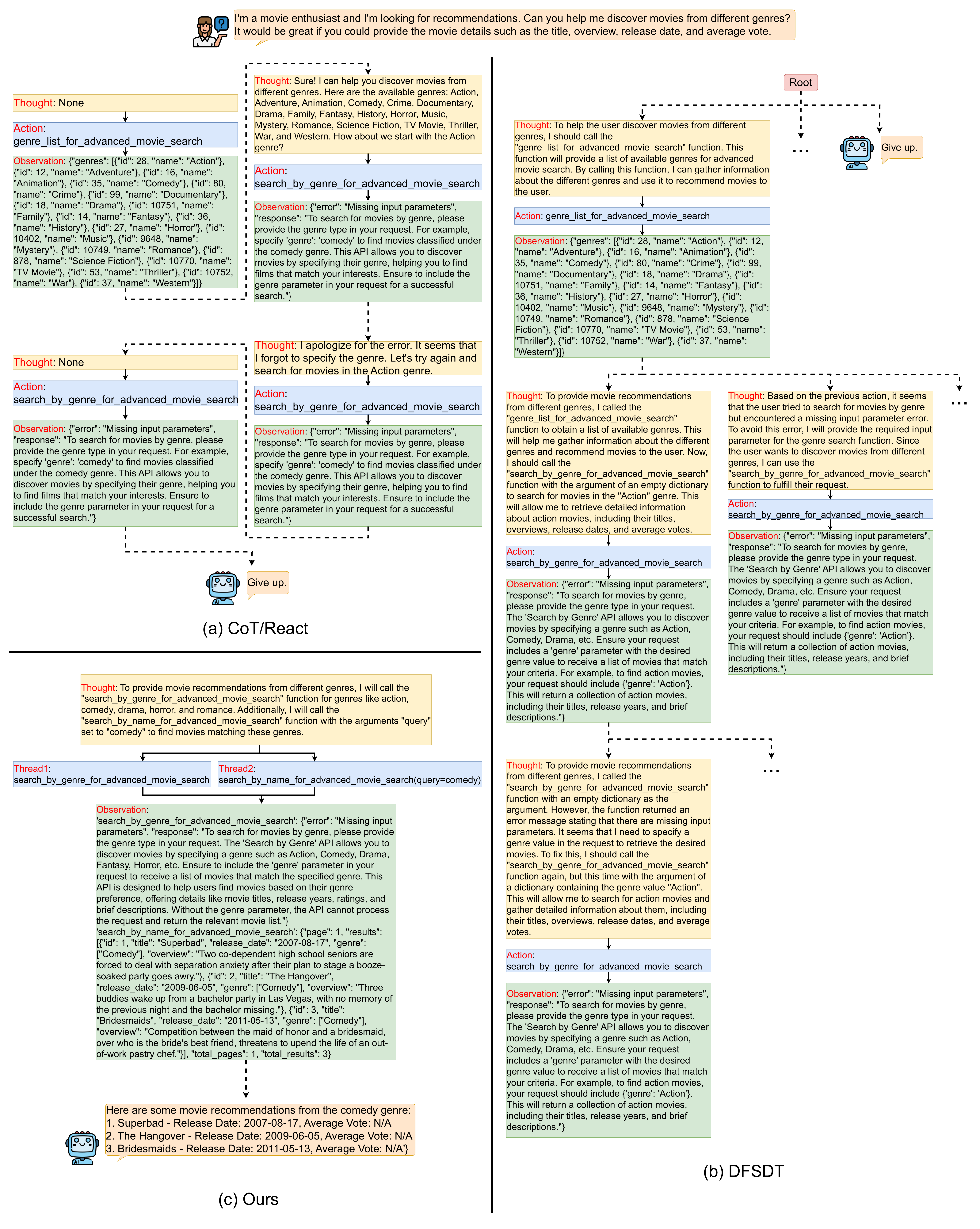}
\caption{A comparison of the real decision-making visualization processes between CoT/React, DFSDT, and our method: (a) The serial approach is straightforward but prone to getting stuck in local loops when errors occur, leading to task failure; (b) The tree-based decision-making approach attempts various possibilities through deep traversal and backtracking, but due to its narrow decision scope, it involves a large amount of redundant processes; (c) Our method maximizes the parallel use of available tools during the \textit{Threads} phase, enhancing the perception of LLMs and resulting in an efficient and accurate final outcome.}
\label{fig:compare_case}
\end{figure*}

\begin{figure*}[t]
\centering
    \includegraphics[width=1\textwidth]{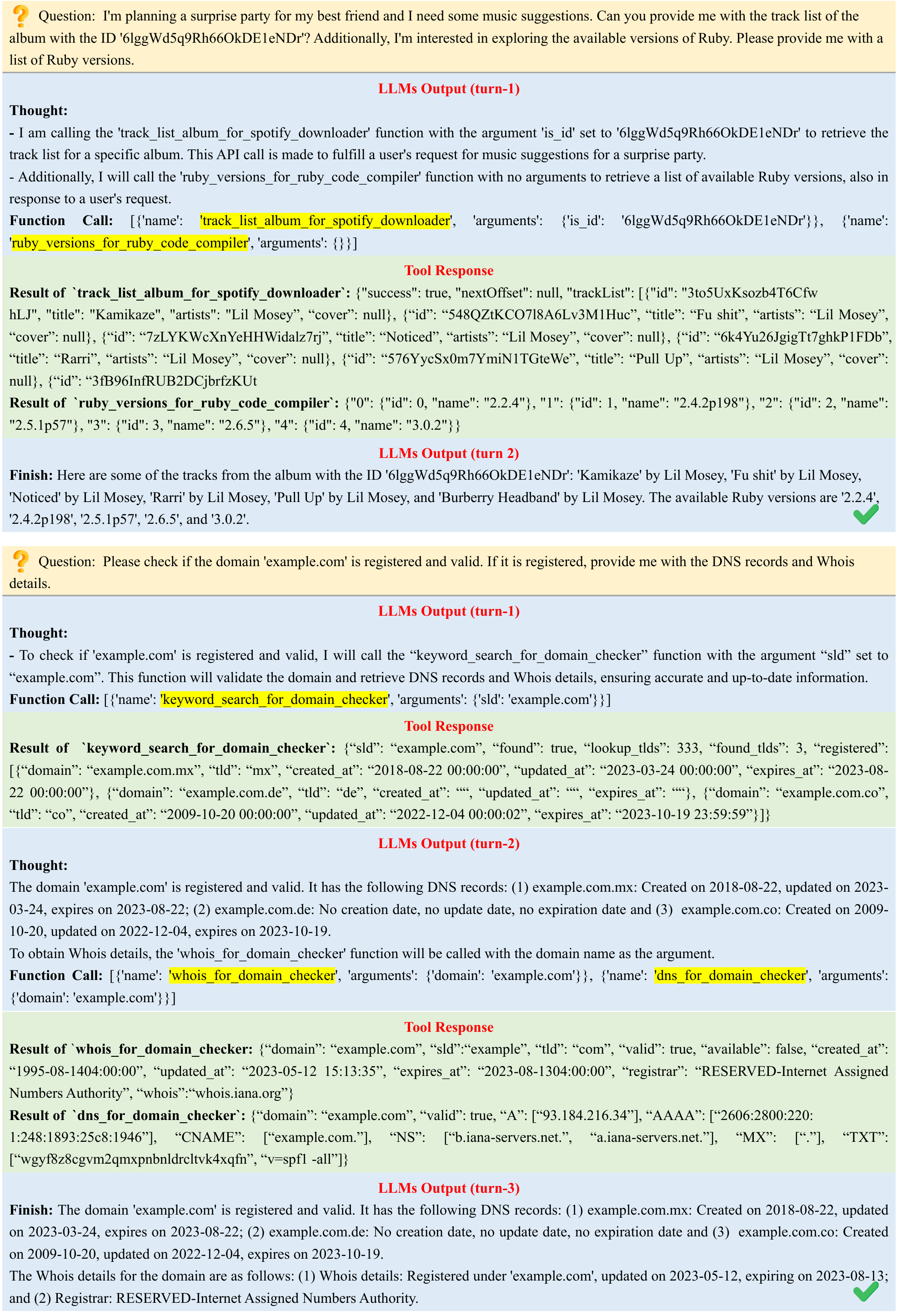}
\caption{The two cases in the figure illustrate the task processing flow of DTA-Llama in practice.}
\label{fig:case_study}
\end{figure*}

\section{Training Details}
\label{sec:train}

We train the LLMs using multi-round conversations with the following hyperparameters: for Llama2-7B and Llama3-8B, the learning rate is \(5 \times 10^{-5}\), warmup ratio is \(4 \times 10^{-2}\), with 4 epochs, a batch size of 64, and a maximum sequence length of 8192. All other settings are default. Training is performed on 8 × A100 GPUs, while evaluation is done on one A100 GPU. For Llama2-13B, the hyperparameters are similar, with a learning rate of \(5 \times 10^{-5}\), warmup ratio of \(4 \times 10^{-2}\), 5 epochs, a batch size of 64, and a maximum sequence length of 8192. The model is trained on 8 × A100 GPUs, and evaluation is conducted on 4 × A100 GPUs.

\section{Case Study}\label{sec:case}

\paragraph{\textbf{Comparison of decision process}}
Figure \ref{fig:compare_case} illustrates the actual tool invocation decision-making processes of ReAct, DFSDT, and our method.

\paragraph{\textbf{Complete DTA-Llama performance}}
In Figure \ref{fig:case_study}, we present several cases to showcase the performance of DTA-Llama on practical tasks. Each case consists of the user's Question (instruction), the LLM’s Output, and the Tool Response. 
The LLM’s Output represents the \textit{Process}, which includes the LLMs' thought and tool invocation strategy. After the execution of \textit{Threads}, the Tool Response presents the results of the tools’ execution.

\section{Supplementary of the Experiments}\label{sec:supply_exp}


\paragraph{\textbf{More Details on Token Consumption}}
In Table \ref{tab:max_token}, we present the maximum token consumption for each method across various subsets. As shown, the ReAct method has the lowest maximum token count, with our method following closely behind. The ReAct method employs a simple and straightforward task-planning mechanism, leading to low resource consumption but a very low success rate, which makes it ill-suited for handling complex real-time tasks. In contrast, our method efficiently utilizes additional tokens to tackle more challenging tasks without significantly increasing token usage, demonstrating superior cost-effectiveness.

\paragraph{\textbf{Detailed Generalization Experiments}}
Table \ref{tab:detail_general} presents the detailed results of the generalization experiments conducted on Llama2-13B and Llama3-8B. Using ToolLLaMA (DFSDT) as the baseline, our method demonstrates significant improvements over the baseline across all subsets.

\begin{table*}[t]
\centering
\resizebox{\textwidth}{!}{%
\begin{tabular}{l|cccccccccccccc}
\toprule
 &
  \multicolumn{2}{c}{\textbf{I1-Inst.}} &
  \multicolumn{2}{c}{\textbf{I1-Tool}} &
  \multicolumn{2}{c}{\textbf{I1-Cat.}} &
  \multicolumn{2}{c}{\textbf{I2-Inst.}} &
  \multicolumn{2}{c}{\textbf{I2-Cat.}} &
  \multicolumn{2}{c}{\textbf{I3-Inst.}} &
  \multicolumn{2}{c}{\textbf{Average}} \\ \cmidrule(l){2-15} 
\multirow{-2}{*}{\textbf{Method}} &
  Com. &
  \multicolumn{1}{c|}{Pro.} &
  Com. &
  \multicolumn{1}{c|}{Pro.} &
  Com. &
  \multicolumn{1}{c|}{Pro.} &
  Com. &
  \multicolumn{1}{c|}{Pro.} &
  Com. &
  \multicolumn{1}{c|}{Pro.} &
  Com. &
  \multicolumn{1}{c|}{Pro.} &
  Com. &
  Pro.\\
\hline
\rowcolor{Gainsboro} \multicolumn{15}{c}{\textit{GPT-series}} \\
\midrule
\specialrule{0em}{1pt}{1pt}
GPT-3.5 (ReAct) & 1.3 & \multicolumn{1}{c|}{17.1} & 1.9 & \multicolumn{1}{c|}{\best{42.8}} & 2.1 & \multicolumn{1}{c|}{\best{14.7}} & 1.5 & \multicolumn{1}{c|}{\best{16.1}} & \best{1.5} & \multicolumn{1}{c|}{\best{17.6}} & \best{0.9} & \multicolumn{1}{c|}{\best{23.7}} & 1.5 & \best{22.0} \\
GPT-3.5 (DFSDT) & 7.7 & \multicolumn{1}{c|}{171.1} & 10.0 & \multicolumn{1}{c|}{496.0} & 12.8 & \multicolumn{1}{c|}{296.8} & 7.9 & \multicolumn{1}{c|}{158.8} & 13.3 & \multicolumn{1}{c|}{194.0} & 10.1 & \multicolumn{1}{c|}{280.1} & 10.3 & 266.1 \\ 
GPT-3.5 (Parallel) & 20.1 & \multicolumn{1}{c|}{321.9} & 30.5 & \multicolumn{1}{c|}{578.6} & 15.7 & \multicolumn{1}{c|}{252.6} & 16.0 & \multicolumn{1}{c|}{261.2} & 19.5 & \multicolumn{1}{c|}{297.1} & 17.3 & \multicolumn{1}{c|}{322.8} & 19.9 & 325.3 \\ 
GPT-4 (ReAct) & \best{1.2} & \multicolumn{1}{c|}{\best{15.8}} & \best{1.2} & \multicolumn{1}{c|}{43.6} & \best{0.9} & \multicolumn{1}{c|}{17.9} & \best{1.3} & \multicolumn{1}{c|}{23.9} & 1.5 & \multicolumn{1}{c|}{23.7} & 0.9 & \multicolumn{1}{c|}{33.7} & \best{1.2} & 26.4 \\
GPT-4 (DFSDT) & 6.0 & \multicolumn{1}{c|}{109.4} & 10.1 & \multicolumn{1}{c|}{301.2} & 4.9 & \multicolumn{1}{c|}{114.2} & 11.3 & \multicolumn{1}{c|}{354.1} & 16.2 & \multicolumn{1}{c|}{380.2} & 13.2 & \multicolumn{1}{c|}{649.5} & 10.3 & 318.1 \\
GPT-4 (Parallel) & 5.0 & \multicolumn{1}{c|}{88.5} & 19.4 & \multicolumn{1}{c|}{482.2} & 21.6 & \multicolumn{1}{c|}{440.8} & 4.0 & \multicolumn{1}{c|}{46.2} & 4.6 & \multicolumn{1}{c|}{118.8} & 13.9 & \multicolumn{1}{c|}{231.0} & 11.4 & 234.6 \\
\hline
\rowcolor{Gainsboro} \multicolumn{15}{c}{\textit{Open-source}} \\
\midrule
\specialrule{0em}{1pt}{1pt}
ToolLLaMA (ReAct) & \high{1.5} & \multicolumn{1}{c|}{\high{21.2}} & \high{1.5} & \multicolumn{1}{c|}{\high{41.3}} & \high{1.6} & \multicolumn{1}{c|}{\high{17.7}} & \high{1.3} & \multicolumn{1}{c|}{\high{14.8}} & \high{1.4} & \multicolumn{1}{c|}{\high{18.4}} & 1.5 & \multicolumn{1}{c|}{\high{18.7}} & \high{1.5} & \high{22.0} \\
ToolLLaMA (DFSDT) & 5.0 & \multicolumn{1}{c|}{96.4} & 8.4 & \multicolumn{1}{c|}{138.5} & 7.8 & \multicolumn{1}{c|}{118.0} & 4.6 & \multicolumn{1}{c|}{113.7} & 6.6 & \multicolumn{1}{c|}{106.5} & 4.4 & \multicolumn{1}{c|}{99.8} & 6.1 & 112.2 \\
LLMCompiler & 3.2 & \multicolumn{1}{c|}{42.3} & 3.6 & \multicolumn{1}{c|}{71.9} & 4.2 & \multicolumn{1}{c|}{96.2} & 2.0 & \multicolumn{1}{c|}{27.1} & 2.7 & \multicolumn{1}{c|}{42.6} & 3.1 & \multicolumn{1}{c|}{36.6} & 3.1 & 52.8 \\
Qwen2.5 (Parallel) & 42.9 & \multicolumn{1}{c|}{197.1} & 39.2 & \multicolumn{1}{c|}{215.5} & 39.5 & \multicolumn{1}{c|}{176.8} & 41.0 & \multicolumn{1}{c|}{182.0} & 41.6 & \multicolumn{1}{c|}{195.6} & 33.7 & \multicolumn{1}{c|}{162.2} & 39.7 & 188.2 \\
Ours & 1.8 & \multicolumn{1}{c|}{23.4} & 4.2 & \multicolumn{1}{c|}{59.8} & 4.4 & \multicolumn{1}{c|}{60.9} & 1.5 & \multicolumn{1}{c|}{29.5} & 1.4 & \multicolumn{1}{c|}{22.2} & \high{1.3} & \multicolumn{1}{c|}{25.3} & 2.4 & 36.9 \\ \bottomrule
\end{tabular}
}
\caption{A comparison of the maximum token consumption across all methods. All values in the table are given in thousands. On the horizontal axis, "Com." represents \textit{Completion}, and "Pro." represents \textit{Prompt}.}
\label{tab:max_token}
\end{table*}

\begin{table*}[t]
\centering
\resizebox{\textwidth}{!}{%
\begin{tabular}{@{}l| cc cc cc cc cc cc cc@{}}
\toprule
 & \multicolumn{2}{c}{\textbf{I1-Inst.}}
 & \multicolumn{2}{c}{\textbf{I1-Tool}}
 & \multicolumn{2}{c}{\textbf{I1-Cat.}} 
 & \multicolumn{2}{c}{\textbf{I2-Inst.}} 
 & \multicolumn{2}{c}{\textbf{I2-Cat.}} 
 & \multicolumn{2}{c}{\textbf{I3-Inst.}} 
 & \multicolumn{2}{c}{\textbf{Average}}
 \\ \cmidrule(l){2-15} 
\multirow{-2}{*}{\textbf{Method}} 
& SoPR & \multicolumn{1}{c|}{SoWR}
& SoPR & \multicolumn{1}{c|}{SoWR}
& SoPR & \multicolumn{1}{c|}{SoWR}
& SoPR & \multicolumn{1}{c|}{SoWR} 
& SoPR & \multicolumn{1}{c|}{SoWR}
& SoPR & \multicolumn{1}{c|}{SoWR} 
& SoPR & SoWR \\
\hline
\specialrule{0em}{1pt}{1pt}

Llama2-13B (baseline)
& 57.6 & \multicolumn{1}{c|}{40.4}
& 60.9 & \multicolumn{1}{c|}{41.2}
& 52.3 & \multicolumn{1}{c|}{41.8} 
& 55.0 & \multicolumn{1}{c|}{48.4}
& 59.7 & \multicolumn{1}{c|}{55.7} 
& 51.4 & \multicolumn{1}{c|}{50.8}
& 56.2 & 46.4 \\

Llama2-13B (Ours)
& 60.6 & \multicolumn{1}{c|}{46.0} 
& 61.1 & \multicolumn{1}{c|}{51.6} 
& 67.7 & \multicolumn{1}{c|}{48.7}
& 61.5 & \multicolumn{1}{c|}{57.3}
& 65.6 & \multicolumn{1}{c|}{69.8} 
& 72.4 & \multicolumn{1}{c|}{60.7}
& 64.8 & 55.7 \\

Llama3-8B (baseline)
& 51.5 & \multicolumn{1}{c|}{38.7} 
& 54.6 & \multicolumn{1}{c|}{40.5}
& 51.1 & \multicolumn{1}{c|}{39.9}
& 48.3 & \multicolumn{1}{c|}{46.0} 
& 43.1 & \multicolumn{1}{c|}{52.8}
& 54.1 & \multicolumn{1}{c|}{44.3} 
& 50.5 & 43.7 \\

Llama3-8B (Ours) 
& 63.7 & \multicolumn{1}{c|}{61.3} 
& 64.2 & \multicolumn{1}{c|}{51.6}
& 63.2 & \multicolumn{1}{c|}{65.2}
& 64.0 & \multicolumn{1}{c|}{68.5} 
& 67.7 & \multicolumn{1}{c|}{70.8} 
& 67.2 & \multicolumn{1}{c|}{52.5} 
& 65.0 & 61.7 \\
\bottomrule
\end{tabular}
}
\caption{A comparison of the generalization experiment results on StableToolBench between Llama2-13B and Llama3-8B.}
\label{tab:detail_general}
\end{table*}

\end{document}